\documentclass[10pt,twocolumn,letterpaper]{article}

\usepackage[pagenumbers]{cvpr} 

\definecolor{cvprblue}{rgb}{0.21,0.49,0.74}
\usepackage[pagebackref,breaklinks,colorlinks,allcolors=cvprblue]{hyperref}
\usepackage{booktabs,adjustbox,multirow,tabularx,threeparttable,makecell}

\def\paperID{12209} 
\def\confName{CVPR}
\def\confYear{2026}

\title{UHKD: A Unified Framework for Heterogeneous Knowledge Distillation via Frequency-Domain Representations}

\author{
Fengming Yu,
Haiwei Pan,
Kejia Zhang,
Jian Guan,
Haiying Jiang \\
Department of Computer Science, Harbin Engineering University, Harbin, China \\
{\tt\small \{yufengming, panhaiwei, kejiazhang, j.guan, jianghaiying\}@hrbeu.edu.cn}
}

\begin{document}
\maketitle
\begin{abstract}
Knowledge distillation (KD) is an effective model compression technique that transfers knowledge from a high-performance 
teacher to a lightweight student, reducing computational and storage costs while maintaining competitive accuracy. 
However, most existing KD methods are tailored for homogeneous models and perform poorly in heterogeneous settings, 
particularly when intermediate features are involved. 
Semantic discrepancies across architectures hinder effective use of intermediate representations from the teacher model, 
while prior heterogeneous KD studies mainly focus on the logits space, underutilizing rich semantic information in intermediate layers. 
To address this, Unified Heterogeneous Knowledge Distillation (UHKD) is proposed, 
a framework that leverages intermediate features in the frequency domain for cross-architecture transfer. 
Frequency-domain representations are leveraged to capture global semantic knowledge and mitigate representational discrepancies between heterogeneous teacher-student pairs.
Specifically, a Feature Transformation Module (FTM) generates compact frequency-domain representations of teacher features, 
while a learnable Feature Alignment Module (FAM) projects student features and aligns them via multi-level matching. 
Training is guided by a joint objective combining mean squared error on intermediate features with Kullback-Leibler divergence on logits. 
Extensive experiments on CIFAR-100 and ImageNet-1K demonstrate the effectiveness of the proposed approach, 
achieving maximum gains of 5.59\% and 0.83\% over the latest heterogeneous distillation method on the two datasets, respectively.
Code will be released soon.
\end{abstract}    
\section{Introduction}
\label{sec:intro}

Knowledge distillation (KD) has emerged as an effective technique for model compression and acceleration, 
aiming to reduce complexity while retaining as much performance as possible.
This makes KD a practical solution for deploying deep networks in resource-constrained environments.
At the same time, the growing pursuit of accuracy has led to increasingly complex architectures such as 
CNNs \cite{DBLP:conf/cvpr/HeZRS16,DBLP:conf/cvpr/SandlerHZZC18,DBLP:conf/cvpr/0003MWFDX22}, 
vision transformers (ViTs) \cite{DBLP:conf/iclr/DosovitskiyB0WZ21,DBLP:conf/icml/TouvronCDMSJ21,DBLP:conf/iccv/LiuL00W0LG21}, 
and MLPs \cite{DBLP:conf/nips/TolstikhinHKBZU21,DBLP:journals/pami/TouvronBCCEGIJSVJ23}, 
whose heavy computational demands further highlight the need for compression.

KD was first introduced by Hinton et al.\cite{DBLP:journals/corr/HintonVD15} as a technique in which a compact student 
is trained under the guidance of a larger, better-performing teacher. 
Besides ground-truth labels, the teacher provides soft targets as auxiliary supervision, helping the student approximate 
task objectives more effectively. 
This strategy reduces model size and cost, thereby improving inference efficiency while largely preserving accuracy. 
Existing approaches can be broadly categorized into three groups\cite{gou2021knowledge,2024-40694}: 
(1) response-based methods, which transfer inter-class probability distributions via soft targets\cite{DBLP:journals/corr/HintonVD15,DBLP:conf/aaai/YangXQY19,DBLP:conf/iccv/SonNCH21}; 
(2) feature-based methods, which distill intermediate representations to enhance representational capacity\cite{DBLP:journals/corr/RomeroBKCGB14,DBLP:conf/aaai/HeoLY019a,DBLP:conf/cvpr/LinXWYCLW22}; 
and (3) relation-based methods, which transfer structural knowledge among samples or features\cite{DBLP:conf/cvpr/YimJBK17,DBLP:conf/cvpr/ParkKLC19,DBLP:conf/iclr/TianKI20}.

\begin{figure}[htbp]
  \centering
  \includegraphics[width=\columnwidth]{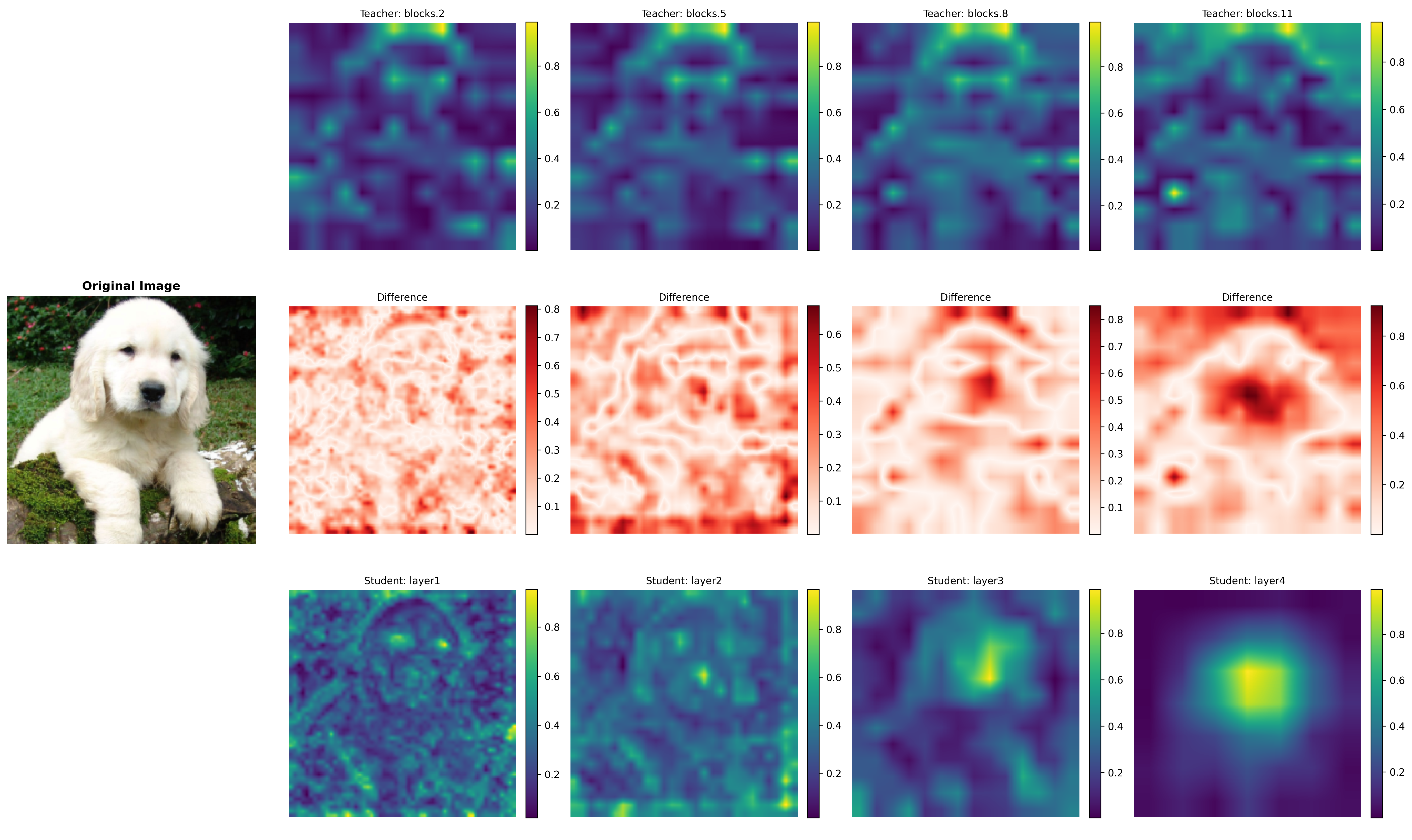} 
  \caption{Intermediate feature visualization of different architectures. \textbf{Left:} original image. 
  \textbf{Top right:} ViT-S (teacher) intermediate feature. 
  \textbf{Bottom right:} ResNet-18 (student) intermediate feature. 
  \textbf{Middle:} difference between teacher and student features.
}
  \label{1-feature-vis}
  \vspace{-0.5cm}
\end{figure}

Although the above methods have achieved promising results in homogeneous distillation, in practice it is often difficult to 
obtain a high-performance teacher that shares the same architecture as the lightweight student. 
In homogeneous settings, intermediate features usually exhibit similar structural patterns, making direct alignment feasible. 
In contrast, heterogeneous models show substantial discrepancies in semantic abstraction and feature distribution, as shown in Fig.~\ref{1-feature-vis}, 
which hinder effective use of intermediate features and lead to suboptimal results when homogeneous methods are directly applied. 
Recent studies on heterogeneous distillation typically focus on fixed transfer directions between specific architecture pairs, 
such as CNN to ViT or ViT to CNN\cite{DBLP:conf/icml/TouvronCDMSJ21,DBLP:conf/cvpr/RenGHXTHZ22,DBLP:conf/iccv/ZhaoSL23,DBLP:conf/accv/LiuCLHDL22,DBLP:conf/mm/ZhaoZHZ023,DBLP:conf/fgr/NiTSD024,DBLP:journals/corr/abs-2404-16386}. 
While these works demonstrate feasibility, they often rely on complex, task-specific designs and remain limited to unidirectional transfer, 
lacking flexibility and generalization to arbitrary architecture pairs.

To overcome the limitations of unidirectional methods, recent studies have proposed more general heterogeneous 
distillation frameworks that handle arbitrary architecture combinations, such as OFA \cite{DBLP:conf/nips/HaoG0TH0X23} 
and FBT \cite{li2025fuse}. 
OFA projects student intermediate representations into the logits space to bypass architectural discrepancies, 
while FBT employs weight sharing to construct auxiliary models that fuse heterogeneous architectures and generate 
additional supervision from logits and penultimate features. 
However, both approaches still rely on the logits space, offering only weak supervision and overlooking the rich 
structural and semantic knowledge in intermediate representations. 
Prior work has shown that neglecting such intermediate features reduces distillation effectiveness \cite{yu2025revisiting},
indicating the need to explicitly leverage intermediate representations and address cross-architecture discrepancies 
for scalable and generalizable frameworks. 

To address these issues, this paper proposes a \textbf{U}nified \textbf{H}eterogeneous \textbf{K}nowledge \textbf{D}istillation (\textbf{UHKD}) 
framework that exploits semantic information in intermediate representations through frequency-domain modeling. 
Frequency features aggregate spatial information and capture global semantic relationships \cite{DBLP:conf/wacv/0007NLP0D24}, 
providing a principled bridge for knowledge transfer across heterogeneous architectures.
Unlike prior fixed-architecture approaches, UHKD is general and applicable to arbitrary model combinations. 
It introduces two key components: a Feature Transformation Module (FTM), which performs spectral refinement and sequence transformation
to obtain compact teacher representations, and a Feature Alignment Module (FAM), which employs a learnable adapter to project 
student features after FFT for better alignment. 
By aligning features in the frequency domain, UHKD alleviates semantic discrepancies and enables more effective knowledge transfer. 
The main contributions of this work are summarized as follows:

\begin{itemize}
  \item UHKD is proposed, a frequency-based framework that enables general and flexible knowledge transfer 
  across arbitrary heterogeneous models by leveraging frequency-domain representations to bridge semantic gaps in intermediate features;
  \item Two key components, the FTM and the FAM, are designed, which together provide an effective mechanism for representing 
  and aligning intermediate features in the frequency domain;
  \item Extensive experiments on CIFAR-100 and ImageNet-1K are conducted, and the results demonstrate that UHKD achieves 
  favorable performance compared to existing baselines under diverse heterogeneous architecture combinations.
\end{itemize}
\section{Related Work}
\subsection{Knowledge Distillation}

Knowledge distillation is a widely used model compression technique that improves lightweight students by 
transferring knowledge from high-capacity teachers. 
Hinton et al.\cite{DBLP:journals/corr/HintonVD15} first introduced KD via soft labels, followed by extensions such as 
response-based\cite{DBLP:conf/cvpr/ZhangXHL18,DBLP:conf/icml/FurlanelloLTIA18,DBLP:conf/aaai/YangXQY19,DBLP:conf/iccv/SonNCH21,Li2021ResKD}, 
feature-based\cite{DBLP:journals/corr/RomeroBKCGB14,DBLP:conf/iclr/ZagoruykoK17,DBLP:conf/aaai/HeoLY019a,DBLP:conf/cvpr/LinXWYCLW22,DBLP:conf/iccv/LaoSLLY23}, 
and relation-based methods\cite{DBLP:conf/cvpr/YimJBK17,DBLP:conf/iccv/PengJLZWLZ019,DBLP:conf/cvpr/ParkKLC19,DBLP:conf/iclr/TianKI20,DBLP:conf/iclr/XuFZXWDX022}.  

Motivated by the success of Transformer architecture, heterogeneous KD has attracted increasing attention. 
Touvron et al.\cite{DBLP:conf/icml/TouvronCDMSJ21} introduced a distillation token for CNN to ViT transfer, 
while Ren et al.\cite{DBLP:conf/cvpr/RenGHXTHZ22} extended this with multiple tokens to capture diverse inductive biases. 
Zhao et al.\cite{DBLP:conf/iccv/ZhaoSL23} decomposed CNN knowledge into local and global components, 
emphasizing local knowledge in early training to exploit inductive biases and introducing global knowledge later to enhance ViT learning. 
Liu et al.\cite{DBLP:conf/accv/LiuCLHDL22} pioneered ViT to CNN distillation by employing cross-attention to bridge and 
align heterogeneous feature representations. 
Other works tackled heterogeneous feature alignment by mapping features into unified receptive-field local representations \cite{DBLP:conf/mm/ZhaoZHZ023}, 
or by enabling collaborative learning among students with diverse inductive biases \cite{DBLP:conf/fgr/NiTSD024}.
Recent frameworks such as OFA\cite{DBLP:conf/nips/HaoG0TH0X23} and FBT\cite{li2025fuse} aim for generality by bridging 
feature gaps through logits or auxiliary models, improving flexibility but still limited in exploiting intermediate semantics.

\subsection{Frequency-Domain Representations}
Frequency-domain features have been widely applied in vision tasks such as 
classification\cite{DBLP:conf/cvpr/0007QSWCR20,DBLP:conf/iclr/WilliamsL18}, 
object detection\cite{DBLP:conf/iccv/00140LZLG23,DBLP:conf/cvpr/ZhongLTKWD22,DBLP:conf/eccv/SunXYXL24,DBLP:journals/pami/LiLAXH13}, 
image generation\cite{DBLP:conf/iccv/JiangDWL21}, and super-resolution\cite{DBLP:conf/eccv/PangLJWLLC20}. 
The amplitude spectrum encodes global attributes such as brightness and texture, 
while the phase spectrum preserves fine-grained structures like edges and orientations\cite{Oppenheim1981}. 
Leveraging these properties, Fourier-based methods have been shown to better capture long-range dependencies and 
enhance texture consistency\cite{DBLP:conf/nips/zhouYHZGLML22,DBLP:journals/corr/abs-2405-18679,DBLP:conf/nips/TancikSMFRSRBN20}, 
leading to more discriminative representations.
Recently, frequency-domain features have also been explored for KD\cite{DBLP:conf/eccv/ShinLLYL22,DBLP:conf/aaai/BinhW22,DBLP:conf/wacv/0007NLP0D24,DBLP:conf/cvpr/ZhangHLJCZ24}. 
For instance, Pham et al.\cite{DBLP:conf/wacv/0007NLP0D24} emphasized that each frequency component aggregates 
information from the entire spatial domain, enabling better modeling of global semantics, 
while Zhang et al.\cite{DBLP:conf/cvpr/ZhangHLJCZ24} proposed suppressing harmful frequency components to alleviate 
information loss during downsampling.

While existing KD methods are mostly constrained to homogeneous settings, aligning intermediate semantics across 
heterogeneous models remains challenging. Frequency-domain features, with their strong global modeling capability, 
provide a promising direction for mitigating semantic discrepancies in heterogeneous architectures.
\section{Method}

\subsection{Overall Framework}

\begin{figure*}[!htbp]
  \centering
  \includegraphics[width=1.0\textwidth]{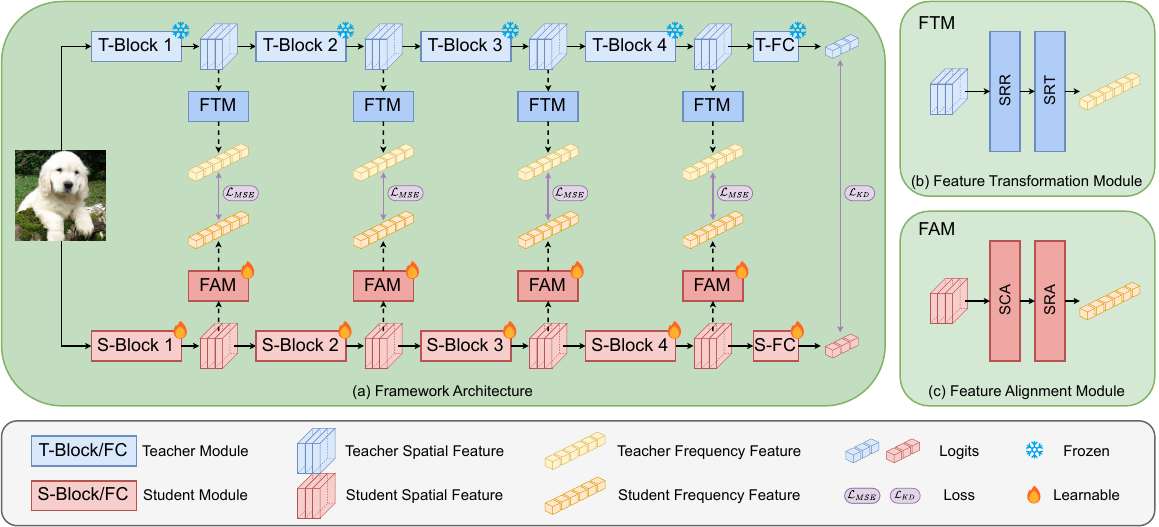} 
  \caption{Overview of unified heterogeneous knowledge distillation. (a) The UHKD framework aligns teacher and student 
  intermediate features in the frequency domain for effective knowledge transfer; 
  (b) FTM module efficiently captures global representations of the teacher model through Spectral Representation 
  Refinement (SRR) and Sequence Representation Transformation (SRT); (c) FAM module adapts student features through Spectral 
  Channel Alignment (SCA) and Sequence Representation Alignment (SRA) to match the frequency-domain features of teacher model.}
  \label{1-Overall}
  \vspace{-0.5cm}
\end{figure*}

This section introduces the Unified Heterogeneous Knowledge Distillation (UHKD) framework, as shown in Fig.~\ref{1-Overall}. 
Direct alignment of intermediate features across heterogeneous architectures is hindered by large distributional gaps. 
UHKD addresses this challenge by transforming teacher and student features into a unified frequency space, 
which ensures consistency in both dimensionality and distribution.
By leveraging semantic knowledge embedded in intermediate representations, UHKD enables flexible and effective 
knowledge transfer across diverse architectures.

\subsection{Feature Transformation Module}
In heterogeneous KD, teacher intermediate features contain diverse semantic information, 
but their spatial distributions and structural forms differ significantly across architectures. 
To enable effective transfer, the FTM converts these features into a unified frequency-domain representation, 
facilitating subsequent alignment and distillation.

\subsubsection{Spectral Representation Refinement}

The Fast Fourier Transform (FFT) is applied to convert teacher features from the spatial domain into the frequency domain. 
FFT is an efficient algorithm that reduces the complexity of the Discrete Fourier Transform \cite{cooley1965algorithm} 
from $\mathcal{O}(N^2)$ to $\mathcal{O}(N \log N)$. After transformation, only the magnitude spectrum is retained:
\begin{equation}
  F^T_{FFT} = \|FFT(F^T)\|_2,
\end{equation}
where $F^T$ denotes teacher features. The magnitude encodes global structural information and energy distribution, 
while the phase mainly captures local details and positional cues \cite{Oppenheim1981, DBLP:conf/iccvw/DongGDC17}. 
For heterogeneous KD, retaining only the magnitude yields stable, architecture-agnostic spectral representations while 
avoiding phase inconsistency that introduces bias across models.

To further refine these spectral features, a frequency filtering mechanism is applied after the FFT. 
The filter modulates components according to their distance from the spectrum center, 
with Gaussian decay ensuring smooth transitions and suppressing boundary artifacts. 
Two complementary masks are constructed: the low-frequency mask emphasizes central components to preserve global structure, 
while the high-frequency mask highlights distant components to retain edges and textures. 
The Gaussian-smoothed mask $M$ is defined as:
\begin{equation}
  M(f) = \exp\!\left(-\left(\frac{||f - f_{c}||}{\sigma}\right)^2\right),
\end{equation}
where $f$ and $f_{c}$ denote the frequency coordinate and spectrum center, respectively, and $\sigma$ is a bandwidth 
parameter controlling the emphasis on low-frequency or high-frequency regions. 
By combining the dual masks, the filter flexibly controls feature distributions across heterogeneous architectures, 
retaining critical information while suppressing redundant and boundary noise. 
The filtered teacher features are then obtained as:
\begin{equation}
  F^T_{filter} = M \odot F^T_{FFT}.
\end{equation}

\subsubsection{Sequence Representation Transformation}




After spectral refinement, a parameter-free downsampling operation transforms teacher representations by removing redundancy and 
emphasizing dominant frequency components, while restructuring them into a unified sequence form $(B, N, C)$ that 
standardizes feature dimensions across heterogeneous teacher models. 
Formally, the transformation is defined as:
\begin{equation}
  F^T_{FTM} \in \mathbb{R}^{B \times N^T \times C^T} = \mathrm{Reshape}(\mathrm{AvgPool}(F^T_{filter})),
\end{equation}
where $F^T_{FTM}$ denotes the output of the FTM, and $N^T$, $C^T$ represent the sequence length and channel dimension of 
the teacher representations. 
This operation yields a compact, structured sequence that reduces computational overhead while preserving essential 
frequency components for effective heterogeneous knowledge transfer.

In summary, FTM enhances spectral features and restructures them into a unified and compact sequence representation, 
enabling efficient heterogeneous knowledge transfer.

\subsection{Feature Alignment Module}
Based on the unified frequency-domain representations of the teacher model, the student features need to be adapted in 
both dimensionality and distribution. 
For this purpose, a learnable FAM is designed, which flexibly projects the intermediate features of the student model into a 
frequency-domain representation consistent with that of the teacher, thereby enabling efficient heterogeneous knowledge transfer.

\subsubsection{Spectral Channel Alignment}


The student features are first transformed into the frequency domain via FFT, 
retaining only the magnitude spectrum as in FTM:
\begin{equation}
  F^S_{FFT} = \|FFT(F^S)\|_2,
\end{equation}
where $F^S$ denotes the intermediate features of the student model. 
To address the dimensionality gap between student and teacher features in the spectral domain, 
a dimension-aware adaptive channel alignment mechanism is introduced. 
It projects the student channels from $C^S$ to $C^T$, ensuring consistency with the teacher representation while preserving structural integrity. 
Formally, the aligned student features are obtained as:
\begin{equation}
  F^S_{CA} = \mathrm{ChannelAlign}_C(F^S_{FFT}),
\end{equation}
where $\mathrm{ChannelAlign}_C$ denotes a learnable channel projection (linear mapping or $1\times1$ convolution). 
This operation adjusts the channel dimensionality while maintaining spatial structure, 
enabling student features from diverse architectures to be efficiently aligned with teacher features.

\subsubsection{Sequence Representation Alignment}

To resolve the mismatch in sequence length between teacher and student features, 
a sequence representation alignment mechanism is applied after channel alignment. 
It reshapes the student features into a compact sequence representation and projects the length from $N^S$ to $N^T$, 
ensuring consistency across heterogeneous architectures. 
Formally, the aligned sequence features are obtained as:
\begin{equation}
 F^S_{SA} \in \mathbb{R}^{B \times N^T \times C^T} = \mathrm{SeqAlign}(F^S_{CA}).
\end{equation}
These features are then normalized to match the distributional statistics of the teacher, 
yielding the final output:
\begin{equation}
  F^S_{FAM} = \mathrm{Norm}(F^S_{SA}),
\end{equation}

In summary, FAM unifies heterogeneous feature representations in the frequency domain with learnable modules, 
yielding a compact and standardized sequence representation consistent with the teacher.

\subsection{Distillation Formulation}

With teacher features transformed by the FTM and student features adapted by the FAM, the heterogeneous intermediate 
representations are mapped into a unified frequency domain for effective alignment and knowledge transfer.

During training, a joint multi-loss strategy is adopted to enhance the student model. 
First, a mean squared error (MSE) loss constrains the consistency of frequency-domain feature distributions:
\begin{equation}
  \mathcal{L}_{MSE} = \frac{1}{B N^T C^T} \left\| F^T_{FTM} - F^S_{FAM} \right\|_2^2.
\end{equation}
Second, a Kullback-Leibler divergence loss aligns the output probability distributions of teacher and student, promoting 
class-level knowledge transfer:
\begin{equation}
  \mathcal{L}_{KL} = \mathrm{D_{KL}}\left( \mathrm{softmax}(z^T / \tau) \,\|\, \mathrm{softmax}(z^S / \tau) \right),
\end{equation}
where $z^T$ and $z^S$ denote the logits of teacher and student, and $\tau$ is the temperature factor. 
Finally, a cross-entropy loss $\mathcal{L}_{CE}$ between student predictions and ground-truth labels preserves 
task-specific discriminative capability.

The three loss terms are combined in a weighted manner to guide student training:
\begin{equation}
  \mathcal{L}_{total} = (1-\lambda_{kl}-\lambda_{ce}) \mathcal{L}_{MSE} + \lambda_{kl} \mathcal{L}_{KL} + \lambda_{ce} \mathcal{L}_{CE},
  \label{eq:total_loss}
\end{equation}
where $\lambda_{kl}$ and $\lambda_{ce}$ are hyperparameters controlling the relative importance of each loss term.
This joint optimization enables the student to learn from multiple perspectives, 
facilitating  effective knowledge transfer and improved generalization in heterogeneous distillation.
\section{Experiment}

\subsection{Experimental Setup}
\subsubsection{Datasets} 

Experiments are conducted on two standard benchmarks, CIFAR-100 \cite{krizhevsky2009learning} and ImageNet-1K \cite{deng2009imagenet}. 
CIFAR-100 consists of 60,000 images of size $32 \times 32$ across 100 classes, with 50,000 for training and 10,000 for testing, 
which poses a fine-grained recognition challenge due to its low resolution and large category set. 
ImageNet-1K contains 1.28 million training and 50,000 validation images in 1,000 classes, with images typically resized to $224 \times 224$, 
and serves as a standard large-scale benchmark for evaluating generalization and scalability.

\subsubsection{Models}

Three representative categories of neural architectures are considered for evaluation. 
CNNs include ResNet \cite{DBLP:conf/cvpr/HeZRS16}, MobileNetV2 \cite{DBLP:conf/cvpr/SandlerHZZC18}, and ConvNeXt \cite{DBLP:conf/cvpr/0003MWFDX22}. 
Transformer-based models cover ViT \cite{DBLP:conf/iclr/DosovitskiyB0WZ21}, DeiT \cite{DBLP:conf/icml/TouvronCDMSJ21}, 
Swin Transformer \cite{DBLP:conf/iccv/LiuL00W0LG21}, and its lightweight variants Swin-Pico and Swin-Nano \cite{DBLP:conf/nips/HaoG0TH0X23}. 
MLP-based models include MLP-Mixer \cite{DBLP:conf/nips/TolstikhinHKBZU21} and ResMLP \cite{DBLP:journals/pami/TouvronBCCEGIJSVJ23}. 
To enable consistent comparison across architectures, all models are uniformly divided into four stages for 
intermediate feature alignment, establishing a common structural granularity for heterogeneous distillation.

\subsubsection{Baselines}

A comprehensive set of representative knowledge distillation methods is included for comparison. 
Feature-based approaches include FitNet \cite{DBLP:journals/corr/RomeroBKCGB14}, CC \cite{DBLP:conf/iccv/PengJLZWLZ019}, RKD \cite{DBLP:conf/cvpr/ParkKLC19}, and CRD \cite{DBLP:conf/iclr/TianKI20}, 
which focus on intermediate representations or relational knowledge. 
Response-based approaches include KD \cite{DBLP:journals/corr/HintonVD15}, DKD \cite{DBLP:conf/cvpr/ZhaoCSQL22}, and DIST \cite{DBLP:conf/nips/0020Y00022}, 
which align output distributions. 
Recent heterogeneous methods such as OFA \cite{DBLP:conf/nips/HaoG0TH0X23} and FBT \cite{li2025fuse} are also considered. 
This selection covers both traditional homogeneous and recent heterogeneous approaches, ensuring a thorough comparison.
Training details are reported in Appendix~\ref{appendix:training_details}.

\subsection{Main Results}

\subsubsection{Results on CIFAR-100} 
Experiments are conducted on 12 heterogeneous teacher-student model pairs with different architectures on CIFAR-100, 
covering CNNs, ViTs, and MLPs. The detailed results are reported in Table~\ref{tab:cifar100}. 

\begin{table*}[ht]
\centering
\footnotesize
\caption{Top-1 accuracy (\%) on CIFAR100. 
The best and second best results are in \textbf{bold} and \underline{underlined}.}
\label{tab:cifar100}
\begin{adjustbox}{max width=\textwidth}
\begin{tabular}{c|cc|cc|cccc|ccc|ccc}
\toprule
\multirow{2}{*}{} & \multirow{2}{*}{Teacher} & \multirow{2}{*}{Student} & \multicolumn{2}{c|}{From Scratch} & \multicolumn{4}{c|}{Feature-based} & \multicolumn{3}{c|}{Response-based} & \multicolumn{3}{c}{Heterogeneous-KD} \\
\cmidrule(r){4-5} \cmidrule(r){6-9} \cmidrule(r){10-12} \cmidrule(r){13-15}
~ & ~ & ~ & T. & S. & FitNet & CC & RKD & CRD & KD & DKD & DIST & OFA & FBT & Ours \\
\midrule

\multirow{6}{*}{\makecell{CNN-based\\students}}
 & Swin-T     & ResNet18    & 89.26 & 74.01 & 78.87 & 74.19 & 74.11 & 77.63 & 78.74 & 80.26 & 77.75 & 80.54 & \underline{81.61} & \textbf{83.13} \\
 & ViT-S      & ResNet18    & 92.04 & 74.01 & 77.71 & 74.26 & 73.72 & 76.60 & 77.26 & 78.10 & 76.49 & 80.15 & \underline{81.93} & \textbf{83.60} \\
 & Mixer-B/16 & ResNet18    & 87.29 & 74.01 & 77.15 & 74.26 & 73.75 & 76.42 & 77.79 & 78.67 & 76.36 & 79.39 & \underline{81.90} & \textbf{82.98} \\
 & Swin-T     & MobileNetv2 & 89.26 & 73.68 & 74.28 & 71.19 & 69.00 & 79.80 & 74.68 & 71.07 & 72.89 & 80.98 & \underline{81.28} & \textbf{83.03} \\
 & ViT-S      & MobileNetv2 & 92.04 & 73.68 & 73.54 & 70.67 & 68.46 & 78.14 & 72.77 & 69.80 & 72.54 & 78.45 & \underline{82.10} & \textbf{84.03} \\
 & Mixer-B/16 & MobileNetv2 & 87.29 & 73.68 & 73.78 & 70.73 & 68.95 & 78.15 & 73.33 & 70.20 & 73.26 & 78.78 & \underline{80.83} & \textbf{83.67} \\
\midrule

\multirow{4}{*}{\makecell{ViT-based\\students}}
 & ConvNeXt-T & DeiT-T      & 88.41 & 68.00 & 60.78 & 68.01 & 69.79 & 65.94 & 72.99 & 74.60 & 73.55 & 75.76 & \textbf{79.57}    & \underline{77.03} \\
 & Mixer-B/16 & DeiT-T      & 87.29 & 68.00 & 71.05 & 68.13 & 69.89 & 65.35 & 71.36 & 73.44 & 71.67 & 73.90 & \underline{74.40} & \textbf{76.26} \\
 & ConvNeXt-T & Swin-P      & 88.41 & 72.63 & 24.06 & 72.63 & 71.73 & 67.09 & 76.44 & 76.80 & 76.41 & 78.32 & \underline{80.73} & \textbf{83.26} \\
 & Mixer-B/16 & Swin-P      & 87.29 & 72.63 & 75.20 & 73.32 & 70.82 & 67.03 & 75.93 & 76.39 & 75.85 & 76.65 & \underline{78.44} & \textbf{81.72} \\
\midrule

\multirow{2}{*}{\makecell{MLP-based\\students}}
 & ConvNeXt-T & ResMLP-S12  & 88.41 & 66.56 & 45.47 & 67.70 & 65.82 & 63.35 & 72.25 & 73.22 & 71.93 & 75.21 & \underline{78.03} & \textbf{83.62} \\
 & Swin-T     & ResMLP-S12  & 89.26 & 66.56 & 63.12 & 68.37 & 64.66 & 61.72 & 71.89 & 72.82 & 11.05 & 73.58 & \underline{77.20} & \textbf{82.72} \\
\midrule

\multicolumn{3}{c|}{Average} & 88.85 & 71.45 & 66.25 & 71.12 & 70.06 & 71.44 & 74.62 & 74.61 & 69.15 & 77.64 & \underline{79.84} & \textbf{82.09} \\
\bottomrule
\end{tabular}
\end{adjustbox}
\end{table*}

Feature-based and response-based distillation methods perform poorly in heterogeneous settings. 
Feature-based approaches average only 69.72\% top-1 accuracy, and direct alignment may even mislead the student, 
as in FitNet where ConvNeXt-T to Swin-P and ConvNeXt-T to ResMLP-S12 yield only 24.06\% and 45.47\%. 
Response-based methods perform slightly better with 72.79\% on average, but still struggle. 
For instance, DIST collapses to 11.05\% when distilling from Swin-T to ResMLP-S12 due to unreliable intra-class relation 
transfer under large feature discrepancies.

Heterogeneous methods such as OFA and FBT achieve competitive results but still make suboptimal use of intermediate representations. 
UHKD aligns teacher and student features in the frequency domain, bridging this gap and reaching 82.09\% average top-1 accuracy, 
surpassing OFA and FBT by 4.45\% and 2.25\%.  
For example, MobileNetv2 students gain up to 5.58\% over OFA (ViT-S to MobileNetv2), while ResMLP-S12 improves by 5.59\% compared to FBT (ConvNeXt-T to ResMLP-S12). 
These results demonstrate the effectiveness of UHKD across diverse heterogeneous architectures over state-of-the-art baselines.

\subsubsection{Results on ImageNet-1K} 

The proposed heterogeneous distillation method UHKD is further evaluated on ImageNet-1K dataset using 12 heterogeneous 
teacher-student model pairs, covering CNNs, ViTs, and MLPs. The detailed results are reported in Table~\ref{tab:imagenet}.

\begin{table*}[!ht]
\centering
\footnotesize
\caption{Top-1 accuracy (\%) on ImageNet-1K. 
The best and second best results are in \textbf{bold} and \underline{underlined}.}
\label{tab:imagenet}
\begin{adjustbox}{max width=\textwidth}
\begin{tabular}{c|cc|cc|cccc|ccc|ccc}
\toprule
\multirow{2}{*}{} & \multirow{2}{*}{Teacher} & \multirow{2}{*}{Student} & \multicolumn{2}{c|}{From Scratch} & \multicolumn{4}{c|}{Feature-based} & \multicolumn{3}{c|}{Response-based} & \multicolumn{3}{c}{Heterogeneous-KD} \\
\cmidrule(r){4-5} \cmidrule(r){6-9} \cmidrule(r){10-12} \cmidrule(r){13-15}
~ & ~ & ~ & T. & S. & FitNet & CC & RKD & CRD & KD & DKD & DIST & OFA & FBT & Ours \\
\midrule

\multirow{6}{*}{\makecell{CNN-based\\students}}
 & DeiT-T     & ResNet18    & 72.17 & 69.75 & 70.44 & 69.77 & 69.47 & 69.25 & 70.22 & 69.39 & 70.64 & 71.01 & \underline{71.22} & \textbf{71.42} \\
 & Swin-T     & ResNet18    & 81.38 & 69.75 & 71.18 & 70.07 & 68.89 & 69.09 & 71.14 & 71.10 & 70.91 & 71.76 & \underline{72.21} & \textbf{72.34} \\
 & Mixer-B/16 & ResNet18    & 76.62 & 69.75 & 70.78 & 70.05 & 69.46 & 68.40 & 70.89 & 69.89 & 70.66 & 71.38 & \underline{71.44} & \textbf{71.45} \\
 & DeiT-T     & MobileNetv2 & 72.17 & 68.87 & 70.95 & 70.69 & 69.72 & 69.60 & 70.87 & 70.14 & 71.08 & 71.39 & \underline{71.78} & \textbf{72.11} \\
 & Swin-T     & MobileNetv2 & 81.38 & 68.87 & 71.75 & 70.69 & 67.52 & 69.58 & 72.05 & 71.71 & 71.76 & 72.32 & \underline{72.54} & \textbf{72.80} \\
 & Mixer-B/16 & MobileNetv2 & 76.62 & 68.87 & 71.59 & 70.79 & 69.86 & 68.89 & 71.92 & 70.93 & 71.74 & 72.12 & \underline{72.31} & \textbf{72.89} \\
\midrule

\multirow{4}{*}{\makecell{ViT-based\\students}}
 & ConvNeXt-T & DeiT-T      & 82.05 & 72.17 & 70.45 & 73.12 & 71.47 & 69.18 & 74.00 & 73.95 & 74.07 & 74.41 & \underline{75.26} & \textbf{76.09} \\
 & Mixer-B/16 & DeiT-T      & 76.62 & 72.17 & 74.38 & 72.82 & 72.24 & 68.23 & 74.16 & 72.82 & 74.22 & 74.46 & \underline{75.00} & \textbf{75.58} \\
 & ConvNeXt-T & Swin-N      & 82.05 & 75.53 & 74.81 & 75.79 & 75.48 & 74.15 & 77.15 & 77.00 & 77.25 & 77.50 & \underline{77.73} & \textbf{77.84} \\
 & Mixer-B/16 & Swin-N      & 76.62 & 75.53 & 76.17 & 75.81 & 75.52 & 73.38 & 76.26 & 75.03 & 76.54 & 76.63 & \underline{76.87} & \textbf{77.26} \\
\midrule

\multirow{2}{*}{\makecell{MLP-based\\students}}
 & ConvNeXt-T & ResMLP-S12  & 82.05 & 76.65 & 74.69 & 75.79 & 75.28 & 73.57 & 76.87 & 77.23 & 77.24 & 77.26 & \underline{77.33} & \textbf{78.05} \\
 & Swin-T     & ResMLP-S12  & 81.38 & 76.65 & 76.48 & 76.15 & 75.10 & 73.40 & 76.67 & 76.99 & 77.25 & 77.31 & \underline{77.42} & \textbf{77.90} \\
\midrule

\multicolumn{3}{c|}{Average} & 78.43 & 72.05 & 72.81 & 72.63 & 71.67 & 70.56 & 73.51 & 73.02 & 73.61 & 73.96 & \underline{74.26} & \textbf{74.64} \\
\bottomrule
\end{tabular}
\end{adjustbox}
\vspace{-0.5cm}
\end{table*}

Feature-based methods achieve 71.92\% average top-1 accuracy, with the severe collapses seen on CIFAR-100 largely 
mitigated by the richer dataset and scale sensitivity of MLPs and Transformers \cite{DBLP:conf/iclr/DosovitskiyB0WZ21,DBLP:conf/nips/TolstikhinHKBZU21,DBLP:conf/iclr/ParkK22}.
Nevertheless, they still struggle with semantic gaps and may even degrade performance. 
For example, FitNet distillation from ConvNeXt-T to ResMLP-S12 achieves 74.69\%, 1.96\% lower than training from scratch. 
Response-based methods perform better, with an average accuracy of 73.38\%, representing a 1.46\% improvement over 
feature-based methods. DIST also benefits from the larger dataset, as the larger number of samples enables more 
reliable inter-class relation transfer.

Heterogeneous methods such as OFA and FBT achieve competitive results, with average top-1 accuracies of 73.96\% and 
74.26\%, corresponding to gains of 0.58\% and 0.88\% over response-based methods. 
In comparison, the proposed UHKD leverages frequency-domain representations and the learnable alignment mechanism in FAM 
to achieve 74.64\%, surpassing OFA and FBT by 0.68\% and 0.38\%, respectively. 
For instance, distillation from ConvNeXt-T to DeiT-T yields 76.09\% top-1 accuracy, 
exceeding OFA by 1.68\% and FBT by 0.83\%. 
Overall, these results demonstrate the robustness and scalability of UHKD across diverse architectures on large-scale datasets.

\subsubsection{Results in Homogeneous Settings on ImageNet-1K}

The proposed UHKD was evaluated on ImageNet-1K under homogeneous settings with two teacher-student pairs, 
ResNet34 to ResNet18 and ResNet50 to MobileNetV2.
Both homogeneous-based \cite{DBLP:conf/iclr/TianKI20,DBLP:conf/cvpr/ZhaoCSQL22,DBLP:conf/nips/0020Y00022,DBLP:conf/cvpr/ChenMZWF022,DBLP:conf/iccv/HeoKYPK019,DBLP:conf/cvpr/Chen0ZJ21,DBLP:conf/iclr/LiuKS023} 
and heterogeneous-based \cite{DBLP:conf/nips/HaoG0TH0X23,li2025fuse} methods were included for comparison. 
As shown in Table~\ref{tab:homo-results-fbt}, UHKD achieves 72.71\% and 73.52\% on the two pairs, surpassing all baselines. 
These results indicate that the proposed UHKD is effective in both heterogeneous and homogeneous scenarios, 
confirming the robustness and general applicability.

\begin{table*}[ht]
\centering
\footnotesize
\caption{Top-1 accuracy (\%) on ImageNet-1K for homogeneous model.
The best and second best results are in \textbf{bold} and \underline{underlined}.}
\label{tab:homo-results-fbt}
\begin{adjustbox}{max width=\textwidth}
\begin{tabular}{c|c|cc|ccccccc|ccc}
\toprule
\multirow{2}{*}{Teacher} & \multirow{2}{*}{Student} & \multicolumn{2}{c|}{From Scratch} & \multicolumn{7}{c|}{Homo. Based} & \multicolumn{3}{c}{Hetero. Based} \\
\cmidrule(r){3-4} \cmidrule(r){5-11} \cmidrule(r){12-14}
~ & ~ & T. & S. & AT & OFD & CRD & Review & DKD & DIST & FCFD & OFA & FBT & Ours \\
\midrule
ResNet34 & ResNet18    & 73.31 & 69.75 & 70.69 & 70.81 & 71.17 & 71.61 & 71.70 & 72.07 & 72.24 & 72.10 & \underline{72.29} & \textbf{72.71} \\
ResNet50 & MobileNetv2 & 79.86 & 68.87 & 69.56 & 71.25 & 71.37 & 72.56 & 72.05 & 73.24 & 73.37 & 73.28 & \underline{73.45} & \textbf{73.52} \\
\bottomrule
\end{tabular}
\end{adjustbox}
\vspace{-0.5cm}
\end{table*}

\subsection{Ablation Study}

\subsubsection{FFT in both FTM and FAM} 
The necessity of the FFT operation in both FTM and FAM is assessed by removing the FFT module and performing feature 
alignment directly in the spatial domain. 
In this setting, teacher features are downsampled and aligned with student features through a learnable adapter. 
As reported in Table~\ref{tab:ablation-fft}(a), the removal of FFT module consistently reduces the performance of 
student models on CIFAR-100. 
These results indicate that large architectural discrepancies between heterogeneous models cannot be effectively bridged 
in the spatial domain, even with a learnable adapter. 
In contrast, the global information captured in the frequency domain allows more effective alignment and leads to 
improved distillation performance.

\subsubsection{Frequency Filter in FTM} 

The importance of the frequency filter in FTM is evaluated by removing it and transferring the full spectrum from 
teacher to student. 
As shown in Table~\ref{tab:ablation-fft}(b), this leads to clear performance degradation because peripheral regions of 
the spectrum contain noise while most discriminative information lies near the center. 
Without filtering, noise is introduced and hampers student learning, whereas the filter suppresses redundant components 
and preserves dominant information. 
The performance drop is even greater than that caused by removing the FFT, indicating that noisy components are 
amplified in the frequency domain and make the student model more vulnerable. 
These results highlight the critical role of frequency filtering in knowledge transfer.

\begin{table}[!htbp]
\centering
\footnotesize
\caption{Ablation study on CIFAR-100: effect of the FFT module (w/o FFT) and frequency filter in FTM (w/o Freq F.).}
\label{tab:ablation-fft}
\begin{adjustbox}{max width=\columnwidth}
\begin{tabular}{@{}cc|ccc@{}}
\toprule
Teacher    & Student     & (a) w/o FFT                & (b) w/o Freq F.        & \textbf{Ours} \\
\midrule
ConvNeXt-T & Swin-P      & 81.95 ({\color{red}-1.31}) & 81.90 ({\color{red}-1.36}) & \textbf{83.26} \\
Mixer-B/16 & DeiT-T      & 75.93 ({\color{red}-0.33}) & 75.69 ({\color{red}-0.57}) & \textbf{76.26} \\
Swin-T     & ResNet18    & 81.88 ({\color{red}-1.25}) & 81.58 ({\color{red}-1.55}) & \textbf{83.13} \\
ConvNeXt-T & ResMLP-S12  & 82.24 ({\color{red}-1.38}) & 82.72 ({\color{red}-0.90}) & \textbf{83.62} \\
ViT-S      & MobileNetv2 & 82.24 ({\color{red}-1.52}) & 82.72 ({\color{red}-1.49}) & \textbf{84.04} \\
Swin-T     & MobileNetv2 & 80.69 ({\color{red}-2.34}) & 80.74 ({\color{red}-2.29}) & \textbf{83.03} \\
\bottomrule
\end{tabular}
\end{adjustbox}
\vspace{-0.5cm}
\end{table}

\subsubsection{Downsampling in FTM} 
The role of downsampling in FTM is examined by removing the downsampling module and directly aligning full-resolution 
frequency features. 
As shown in Table~\ref{tab:ablation-ds}, this degrades performance across all evaluated teacher-student pairs. 
The results demonstrate the importance of downsampling, which reduces teacher feature resolution, enabling the student 
model to align more effectively, and aggregates local information while suppressing noise to support generalization, 
also reducing FAM parameter overhead. 

\begin{table}[!htbp]
\centering
\footnotesize
\caption{Ablation study on CIFAR-100: effect of DownSampling in FTM (w/o DownSampling).}
\label{tab:ablation-ds}
\begin{adjustbox}{max width=\columnwidth}
\begin{tabular}{cc|cc}
\toprule
Teacher & Student Pair & w/o DownSampling & \textbf{Ours} \\
\midrule
Swin-T & ResNet18        & 82.04 ({\color{red}-1.09}) & \textbf{83.13} \\ 
Swin-T & MobileNetv2     & 82.88 ({\color{red}-0.15}) & \textbf{83.03} \\
Swin-T & ResMLP-S12      & 82.50 ({\color{red}-0.22}) & \textbf{82.72} \\
ConvNeXt-T & Swin-P      & 81.32 ({\color{red}-1.94}) & \textbf{83.26} \\
ConvNeXt-T & ResMLP-S12  & 83.58 ({\color{red}-0.04}) & \textbf{83.62} \\
\bottomrule
\end{tabular}
\end{adjustbox}
\vspace{-0.5cm}
\end{table}

\subsubsection{Learnable Module in FAM} 

The effectiveness of learnable parameters in FAM is assessed by comparison with non-parametric strategies, 
including bilinear interpolation (Bilinear), nearest-neighbor interpolation (Nearest), and a randomly initialized 
non-trainable variant (Random Init.). 
As shown in Table~\ref{tab:alignment-comparison}, all alternatives degrade performance, 
with interpolation in the ConvNeXt-T to Swin-P leading to an accuracy drop of over 16\%. 
In contrast, the learnable FAM adds only a few parameters yet it more effectively adapts to feature discrepancies, 
achieving a favorable balance between parameter overhead and accuracy. 
These results demonstrate the importance of learnable adaptation for robust feature alignment and knowledge transfer.

\begin{table}[htbp]
\centering
\footnotesize
\caption{Ablation study on CIFAR-100: effect of different alignment strategies in FAM.}
\label{tab:alignment-comparison}
\begin{threeparttable}
\begin{adjustbox}{max width=\columnwidth}
\begin{tabular}{c|ccc}
\toprule
\multirow{2}{*}{Method} & T: ViT-S                               & T: ConvNeXt-T               & T: Swin-T \\
                        & S: ResNet18                            & S: Swin-P                   & S: ResMLP-S12 \\
\midrule        
Bilinear                & 82.58$^{\dagger}$ ({\color{red}-1.02}) & 66.75 ({\color{red}-16.51}) & 79.91 ({\color{red}-2.81}) \\
Nearest                 & 82.55 ({\color{red}-1.05})             & 67.22 ({\color{red}-16.04}) & 77.85 ({\color{red}-4.87}) \\
Random Init.            & 82.03 ({\color{red}-1.57})             & 78.92 ({\color{red}-4.34})  & 76.89 ({\color{red}-5.83}) \\
\textbf{Ours}           & \textbf{83.60}                         & \textbf{83.26}              & \textbf{82.72} \\
\bottomrule
\end{tabular}
\end{adjustbox}
\begin{tablenotes}[flushleft]
\footnotesize
\raggedright
\item[$\dagger$] denotes linear interpolation for ViT-based teachers.
\end{tablenotes}
\end{threeparttable}
\vspace{-0.5cm}
\end{table}

\subsubsection{FTM/FAM Branch Counts and Layer Positions} 

The impact of the number and placement of distillation branches is examined on two teacher-student pairs with 1-4 branches, 
as reported in Table~\ref{tab:stage-ablation}. 
Using only one branch significantly reduces performance regardless of its position. With two or three branches, 
deeper layers, especially the final one, provide greater benefits and narrow the gap with the default four-branch setting. 
Since deeper features capture global semantics and shallower ones provide complementary local details, 
combining branches across shallow, intermediate, and deep layers allows both types of information to be exploited, 
making the four-branch configuration the most effective.

\begin{table}[htbp]
\centering
\caption{Ablation study on CIFAR-100: effect of different distillation branch counts and layer positions in FTM/FAM.}
\label{tab:stage-ablation}
\begin{adjustbox}{max width=\columnwidth}
\begin{tabular}{@{}c|cc@{}}
\toprule
Stage & \textbf{{T:}} ConvNeXt-T~~\textbf{S:} Swin-P & \textbf{T:} ViT-S~~\textbf{S:} ResNet18 \\
\midrule              
\{1\}                  & 82.01 ({\color{red}-1.25}) & 83.10 ({\color{red}-0.50}) \\
\{2\}                  & 82.74 ({\color{red}-0.52}) & 83.05 ({\color{red}-0.55}) \\
\{3\}                  & 82.55 ({\color{red}-0.71}) & 82.79 ({\color{red}-0.81}) \\
\{4\}                  & 82.50 ({\color{red}-0.76}) & 83.03 ({\color{red}-0.57}) \\
\{1, 2\}               & 83.19 ({\color{red}-0.07}) & 83.38 ({\color{red}-0.22}) \\
\{2, 3\}               & 82.65 ({\color{red}-0.61}) & 82.93 ({\color{red}-0.67}) \\
\{3, 4\}               & 82.96 ({\color{red}-0.30}) & 83.53 ({\color{red}-0.07}) \\
\{1, 2, 3\}            & 83.02 ({\color{red}-0.24}) & 83.01 ({\color{red}-0.59}) \\
\{2, 3, 4\}            & 83.20 ({\color{red}-0.06}) & 83.54 ({\color{red}-0.06}) \\
\{1, 2, 3, 4\}         & \textbf{83.26}             & \textbf{83.60} \\
\bottomrule
\end{tabular}
\end{adjustbox}
\vspace{-0.5cm}
\end{table}

\subsection{Discussion}

\begin{figure*}[htbp]
  \centering
  \includegraphics[width=\textwidth]{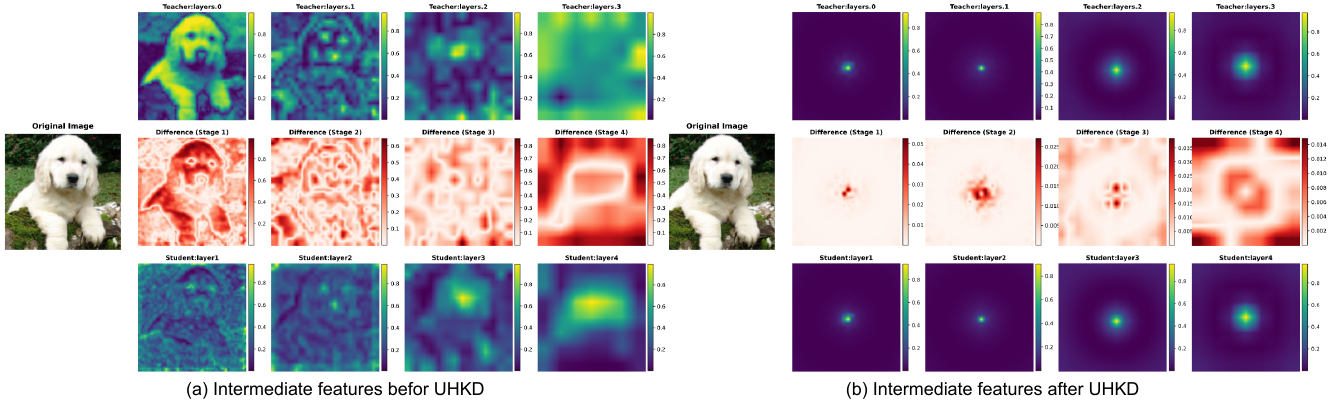} 
  \caption{Visualization of intermediate features before and after UHKD. 
  \textbf{(a)} Before UHKD; \textbf{(b)} After UHKD. 
  In each case, the left column shows the original image, the top and bottom rows show feature maps from different stages 
  of the Swin-T teacher and ResNet-18 student, and the middle row shows their difference maps.}
  \label{3-vis-comparison}
  \vspace{-0.2cm}
\end{figure*}

\begin{table*}[t]
\centering
\caption{Comparison of parameter analysis between OFA and UHKD.}
\label{tab:parameter-analysis}
\begin{adjustbox}{max width=\textwidth}
\scriptsize
\begin{tabular}{@{}cc|cc|ccccc|ccccc@{}}
\toprule 
\multirow{4}{*}{Teacher} & 
\multirow{4}{*}{Student} & 
\multirow{4}{*}{\makecell{Teacher\\Para. (M)}} & 
\multirow{4}{*}{\makecell{Student\\Para. (M)}} & 
\multicolumn{5}{c|}{OFA} & 
\multicolumn{5}{c}{UHKD} \\
\cmidrule(lr){5-9} \cmidrule(lr){10-14}
& & & & 
\makecell{Extra\\(M)} & \makecell{Trainable\\(M)} & \makecell{Extra\\Ratio (\%)} & \makecell{Inference\\(M)} & \makecell{Comp.\\Ratio (\%)} & 
\makecell{Extra\\(M)} & \makecell{Trainable\\(M)} & \makecell{Extra\\Ratio (\%)} & \makecell{Inference\\(M)} & \makecell{Comp.\\Ratio (\%)} \\
\midrule

DeiT-T     & ResNet18    & 5.72  & 11.69 & 3.04 & 14.73 & 20.64 & 11.69 & 204.37 & \textbf{2.01} & \textbf{13.70} & \textbf{14.67} & 11.69 & 204.37 \\
Swin-T     & ResNet18    & 28.33 & 11.69 & \textbf{4.39} & \textbf{16.08} & \textbf{27.30} & 11.69 & 41.26 & 7.28 & 18.97 & 38.38 & 11.69 & 41.26 \\
Mixer-B/16 & ResNet18    & 59.88 & 11.69 & 4.39 & 16.08 & 27.30 & 11.69 & 19.52 & \textbf{3.12} & \textbf{14.81} & \textbf{21.07} & 11.69 & 19.52 \\
DeiT-T     & MobileNetV2 & 5.72  & 3.50  & 7.14 & 10.64 & 67.11 & 3.50  & 61.19 & \textbf{1.83} & \textbf{5.33 } & \textbf{34.33} & 3.50  & 61.19 \\
Swin-T     & MobileNetV2 & 28.33 & 3.50  & 7.14 & 10.64 & 67.11 & 3.50  & 12.35 & \textbf{6.82} & \textbf{10.32} & \textbf{66.09} & 3.50  & 12.35 \\
Mixer-B/16 & MobileNetV2 & 59.88 & 3.50  & 7.14 & 10.64 & 67.11 & 3.50  & 5.85  & \textbf{2.37} & \textbf{5.87 } & \textbf{40.37} & 3.50  & 5.85 \\
\midrule

ConvNeXt-T & DeiT-T & 28.59 & 5.72 & 4.48 & 10.20 & 43.92 & 5.72 & 20.01 & \textbf{1.28} & \textbf{7.00 } & \textbf{18.29} & 5.72 & 20.01 \\
Mixer-B/16 & DeiT-T & 59.88 & 5.72 & 4.48 & 10.20 & 43.92 & 5.72 & 9.55  & \textbf{1.50} & \textbf{7.22 } & \textbf{20.78} & 5.72 & 9.55 \\
ConvNeXt-T & Swin-N & 28.59 & 9.60 & 28.19 & 37.79 & 74.60 & 9.60 & 33.58 & \textbf{6.36} & \textbf{15.96} & \textbf{39.85} & 9.60 & 33.58 \\
Mixer-B/16 & Swin-N & 59.88 & 9.60 & 28.19 & 37.79 & 74.60 & 9.60 & 16.03 & \textbf{3.12} & \textbf{12.72} & \textbf{24.53} & 9.60 & 16.03 \\
\midrule

ConvNeXt-T & ResMLP-S12 & 28.59 & 15.35 & 16.32 & 31.67 & 51.53 & 15.35 & 53.69 & \textbf{1.59} & \textbf{16.94} & \textbf{9.39} & 15.35 & 53.69 \\
Swin-T     & ResMLP-S12 & 28.33 & 15.35 & 16.32 & 31.67 & 51.53 & 15.35 & 54.18 & \textbf{1.59} & \textbf{16.94} & \textbf{9.39} & 15.35 & 54.18 \\
\bottomrule
\end{tabular}
\end{adjustbox}
\vspace{-0.5cm}
\end{table*}

To better understand the role of UHKD in mitigating heterogeneous representation discrepancy, a visual analysis is provided. 
As shown in Fig.~\ref{3-vis-comparison}(a), heterogeneous models (Swin-T and ResNet-18) exhibit substantial 
differences in feature structures and activation distributions, which hinder transfer under direct spatial matching. 
After applying FTM and FAM, the transformed features in Fig.~\ref{3-vis-comparison}(b) display a more compact spectral 
distribution with energy concentrated near the center, reducing structural discrepancies and indicating that the frequency 
domain serves as a normalization space for alignment across architectures. Additional statistical and visual analyses are reported in Appendix~\ref{appendix:supplementary_analyses}.

A parameter analysis comparing UHKD with OFA is presented in Table~\ref{tab:parameter-analysis}, 
which reports the number of extra parameters (Extra), trainable parameters during distillation (Trainable), 
the proportion of extra parameters (Extra Ratio), inference parameters (Inference), and the compression ratio (Comp. Ratio). 
UHKD employs simple components for unified alignment across heterogeneous architectures and applies 
downsampling technology to teacher features, thereby avoiding complex adapters and reducing the parameter demand of the FAM. 
As a result, UHKD consistently requires fewer additional parameters than OFA. 
For instance, UHKD reduces extra parameters from 28.19M to 3.12M (74.60\% to 24.53\%) in Mixer-B/16 to Swin-N, and from 16.32M to 1.59M (51.53\% to 9.39\%) in ConvNeXt-T to ResMLP-S12.
These reductions demonstrate that alignment and effective knowledge transfer can be achieved across diverse 
architectures with relatively low overhead.
\section{Conclusion}

In this paper, a unified heterogeneous knowledge distillation framework, \textbf{UHKD}, is proposed, 
introducing the frequency domain as a bridge for transferring intermediate representations across diverse architectures. 
Leveraging FFT, the FTM compacts and enhances teacher features, while the FAM adapts student features for robust 
cross-architecture alignment. 
In this way, UHKD overcomes the limitations of prior approaches that are restricted to homogeneous settings and rely 
mainly on logits-based supervision, leading to suboptimal use of intermediate semantic information. 

Extensive experiments on CIFAR-100 and ImageNet-1K demonstrate that UHKD consistently outperforms existing methods 
across diverse models. These results highlight the effectiveness of frequency-domain representations in 
capturing global semantics and establish UHKD as a robust and unified framework for heterogeneous knowledge transfer,
offering new insights into intermediate feature alignment and promoting model compression for efficient deployment.

{
    \small
    \bibliographystyle{ieeenat_fullname}
    \bibliography{main}
}

\clearpage
\setcounter{page}{1}
\maketitlesupplementary
\appendix

\section{Training Details} 
\label{appendix:training_details}
All models are trained using the AdamW\cite{DBLP:conf/iclr/LoshchilovH19} optimizer with momentum parameters
$(\beta_1=0.9,\ \beta_2=0.999)$, a weight decay of $0.005$, and a numerical stability constant $\epsilon = 1 \times 10^{-8}$.
A cosine learning rate schedule with warm-up is adopted to facilitate stable convergence. To further regularize training, 
label smoothing with a factor of $0.1$ is applied and gradient clipping with a maximum norm of $5.0$ is employed to prevent exploding gradients.
For data augmentation, a strong strategy is used that combines RandAugment\cite{DBLP:conf/nips/CubukZS020}, 
Mixup\cite{DBLP:conf/iclr/ZhangCDL18}, CutMix\cite{DBLP:conf/iccv/YunHCOYC19}, and random erasing\cite{DBLP:conf/aaai/Zhong0KL020}, 
in addition to standard techniques such as color jittering, random cropping, and horizontal flipping.
The total loss follows the formulation in Eq.~\ref{eq:total_loss}, where $\lambda_{kl}=0.4$ and $\lambda_{ce}=0.3$, resulting in 
relative weights of $0.3$, $0.4$, and $0.3$ for the mean squared error, Kullback-Leibler divergence, and cross-entropy 
terms, respectively. 
For feature-level distillation, four alignment points are selected uniformly along the network depth to capture shallow, 
middle, and deep stages of representation learning.
All experiments are trained until convergence under carefully controlled settings, where optimization and augmentation 
strategies follow the same overall design.

\section{Supplementary Analyses of UHKD}
\label{appendix:supplementary_analyses}

\begin{figure}[htbp]
  \centering
  \includegraphics[width=\columnwidth]{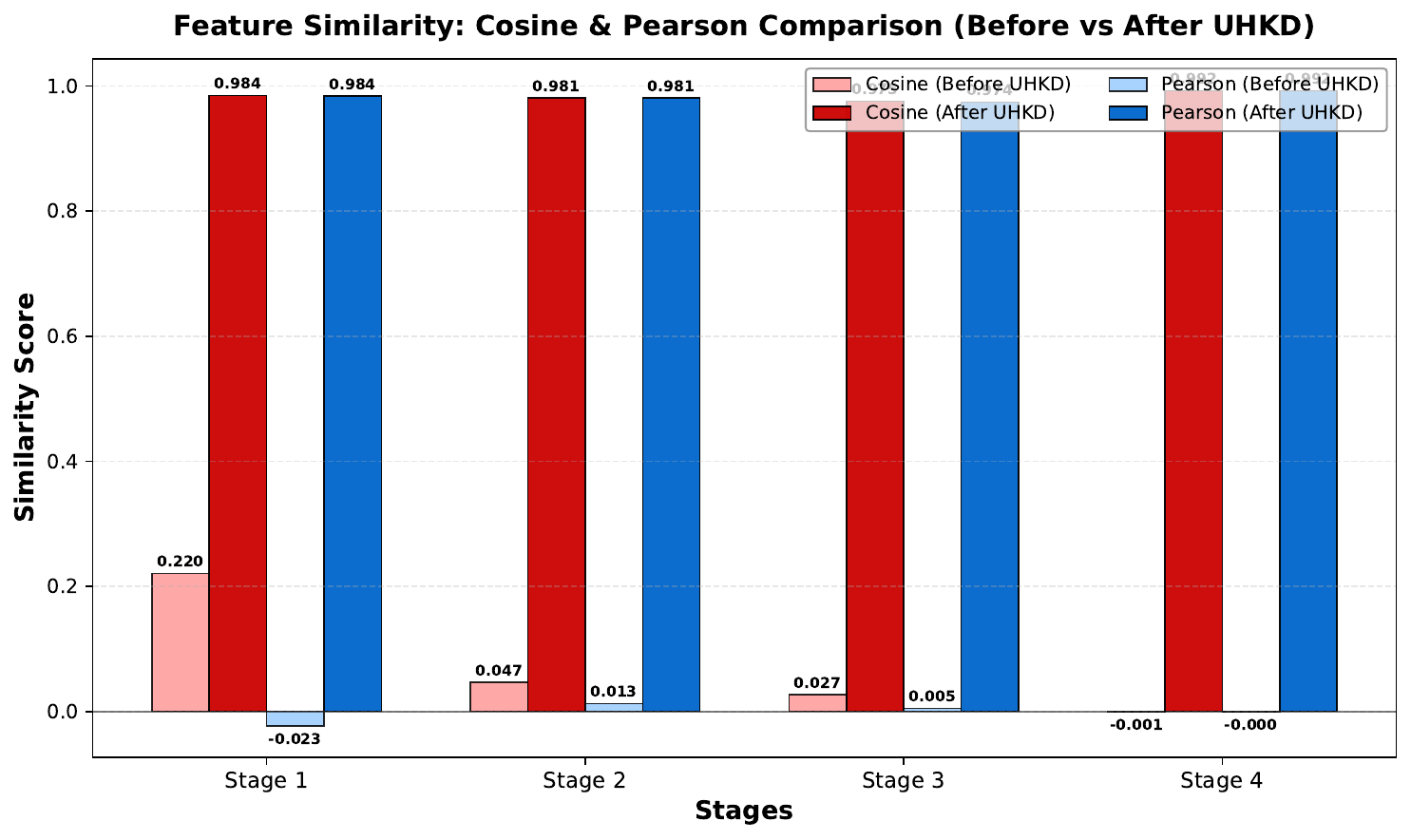} 
  \caption{Comparison of feature similarities before and after UHKD between Swin-T and ResNet-18. 
  Red bars denote cosine similarity, and blue bars denote Pearson correlation.}
  \label{fig:sta-swin-resnet}
\end{figure}

To further illustrate the effectiveness of UHKD, additional visualization and statistical analyses are provided. 
Cosine similarity and Pearson correlation coefficients are calculated between teacher and student feature maps at 
different stages, as shown in Fig.~\ref{fig:sta-swin-resnet}. 
Before UHKD, cosine similarities remain consistently low, with values close to zero in the deep stage, 
indicating a pronounced representational gap. 
After the application of FTM and FAM, similarities increase significantly across all stages, 
with the most notable improvements observed in deeper layers where semantic abstraction is more prominent. 
A comparable trend is observed in the Pearson correlation analysis. 
The coefficients are close to zero before transformation, reflecting uncorrelated feature structures, 
but rise toward the upper bound after UHKD processing, 
signifying strong correspondence and structural coherence between teacher and student representations.

\begin{figure}[htbp]
  \centering
  \includegraphics[width=\columnwidth]{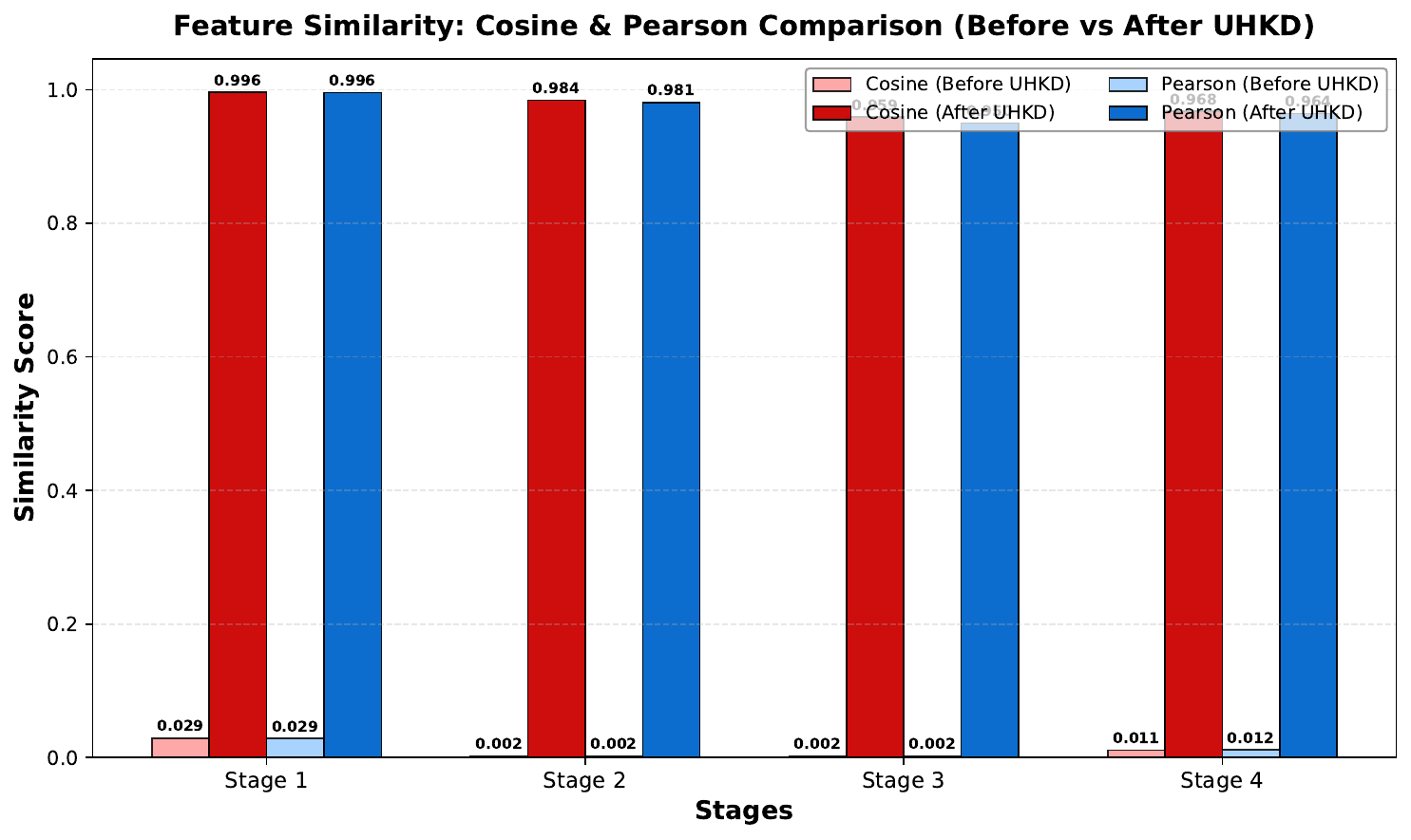} 
  \caption{Comparison of feature similarities before and after UHKD between Swin-T and ResMLP-S12. 
  Red bars denote cosine similarity, and blue bars denote Pearson correlation.}
  \label{fig:sta-swin-resmlp}
  \vspace{-0.5cm}
\end{figure}

\begin{figure*}[htbp]
  \centering
  \includegraphics[width=\textwidth]{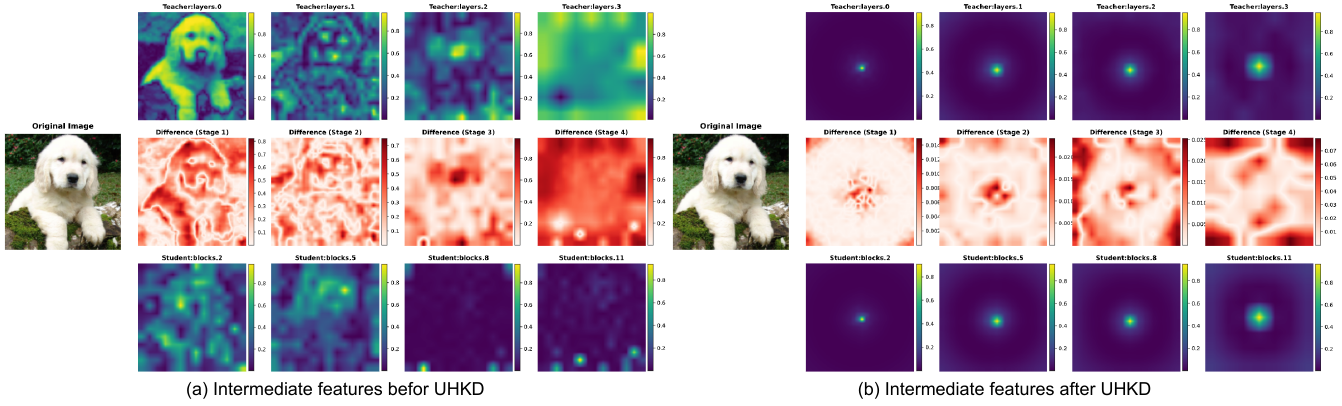} 
  \caption{Visualization of intermediate features before and after UHKD. 
  \textbf{(a)} Before UHKD; \textbf{(b)} After UHKD. 
  In each case, the left column shows the original image, the top and bottom rows show feature maps from different stages 
  of the Swin-T teacher and ResMLP-S12 student, and the middle row shows their difference maps.}
  \label{fig:vis-swin-resmlp}
\end{figure*}

\begin{figure*}[htbp]
  \centering
  \includegraphics[width=\textwidth]{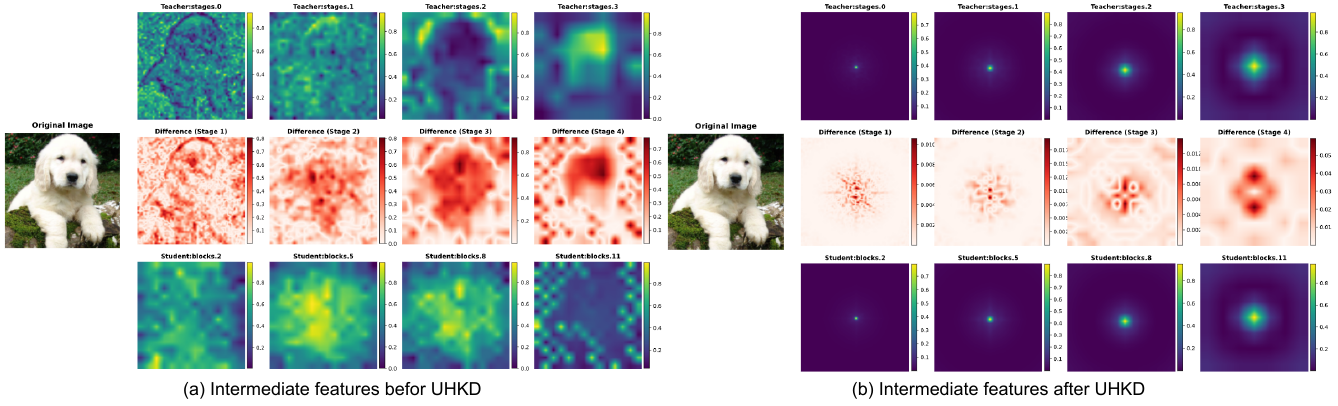} 
  \caption{Visualization of intermediate features before and after UHKD. 
  \textbf{(a)} Before UHKD; \textbf{(b)} After UHKD. 
  In each case, the left column shows the original image, the top and bottom rows show feature maps from different stages 
  of the ConvNeXt-T teacher and DeiT-T student, and the middle row shows their difference maps.}
  \label{fig:vis-convnext-deit} 
\end{figure*}

\begin{figure}[htbp]
  \centering
  \includegraphics[width=\columnwidth]{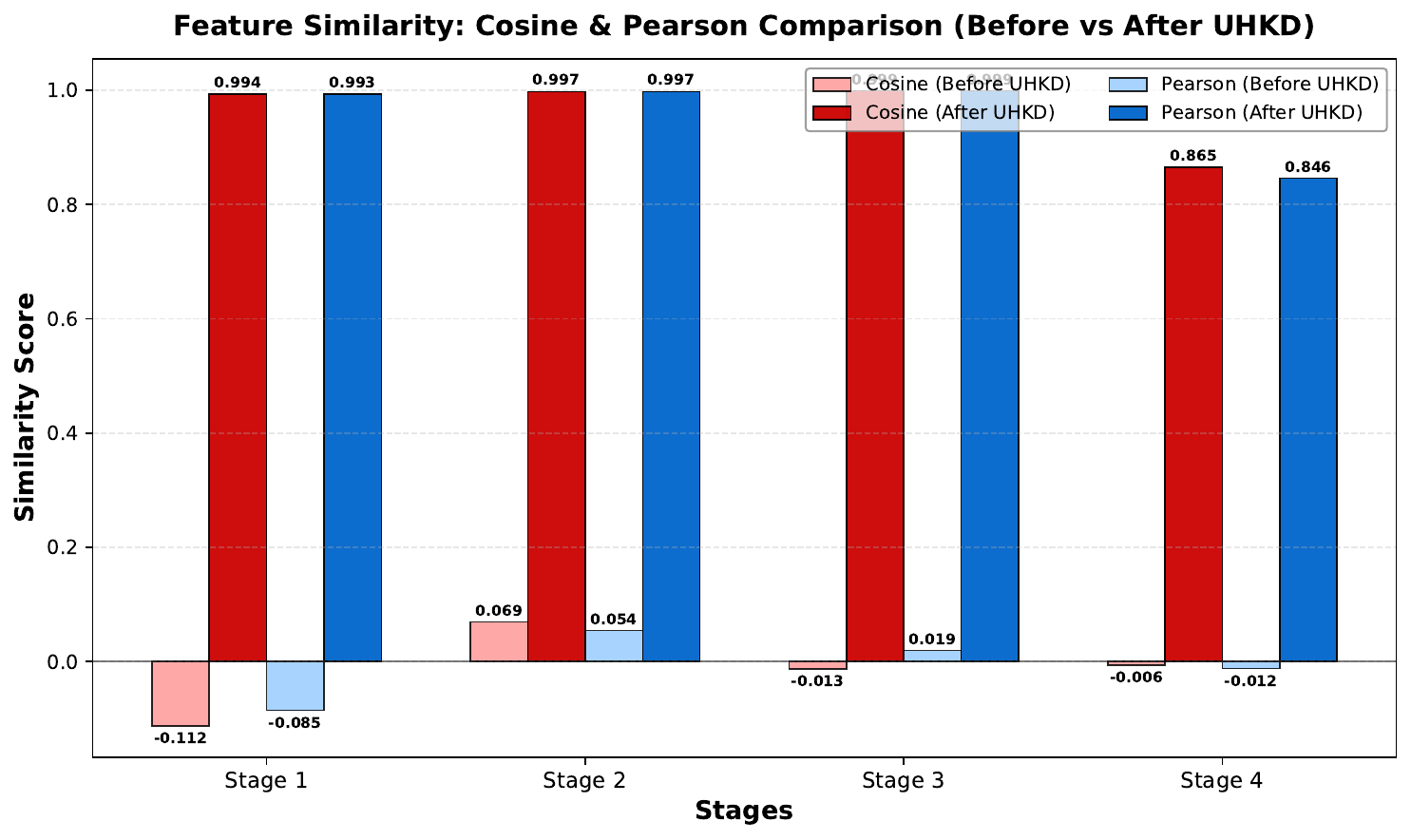} 
  \caption{Comparison of feature similarities before and after UHKD between ConvNeXt-T and DeiT-T. 
  Red bars denote cosine similarity, and blue bars denote Pearson correlation.}
  \label{fig:sta-convnext-deit}
\end{figure}

\begin{figure}[htbp]
  \centering
  \includegraphics[width=\columnwidth]{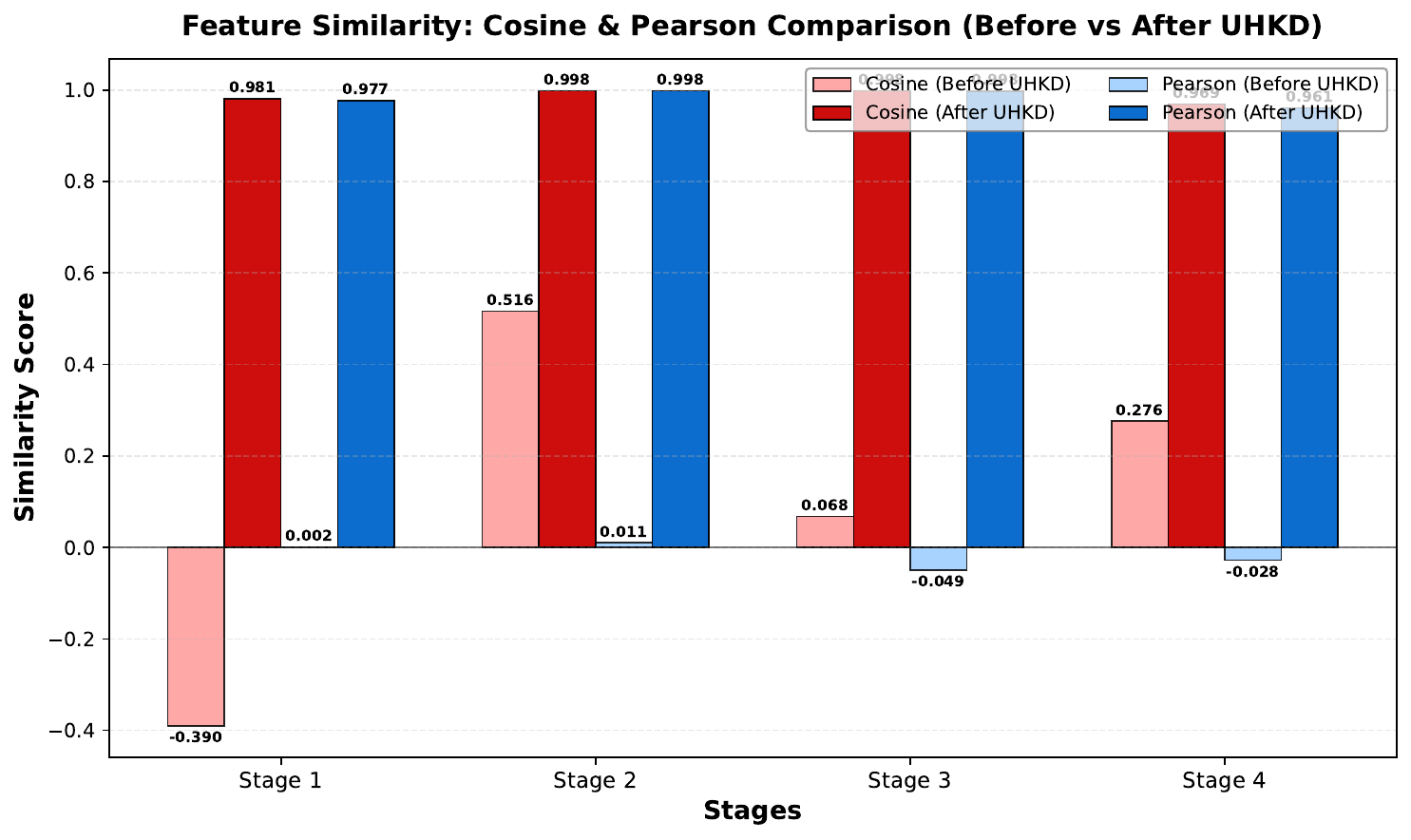} 
  \caption{Comparison of feature similarities before and after UHKD between Mixer-B/16 and Swin-N. 
  Red bars denote cosine similarity, and blue bars denote Pearson correlation.}
  \label{fig:sta-mixer-swin}
\end{figure}

\begin{figure*}[htbp]
  \centering
  \includegraphics[width=\textwidth]{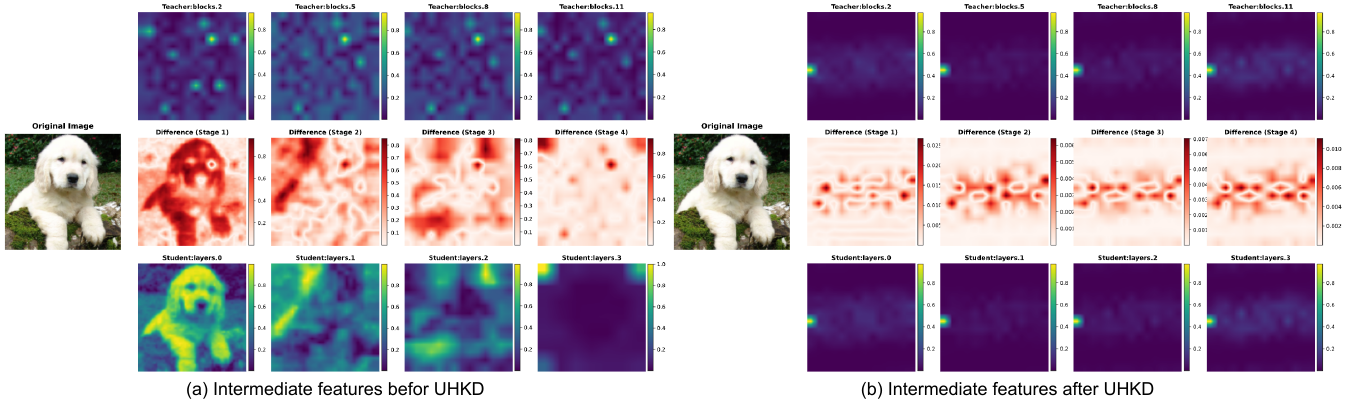} 
  \caption{Visualization of intermediate features before and after UHKD. 
  \textbf{(a)} Before UHKD; \textbf{(b)} After UHKD. 
  In each case, the left column shows the original image, the top and bottom rows show feature maps from different stages 
  of the Mixer-B/16 teacher and Swin-N student, and the middle row shows their difference maps.}
  \label{fig:vis-mixer-swin}
\end{figure*}

Building upon these observations, supplementary analyses are conducted on additional teacher-student pairs. 
The similarity results, as illustrated in Fig.~\ref{fig:sta-swin-resmlp}, \ref{fig:sta-convnext-deit}, and \ref{fig:sta-mixer-swin}, 
show that both cosine similarity and Pearson correlation remain low before UHKD, particularly in deeper layers, 
reflecting substantial representational gaps. 
After the application of FTM and FAM, significant improvements are observed across all stages. 
The corresponding visualizations reveal more compact spectral distributions and reduced structural discrepancies, 
supporting the statistical findings, as shown in Fig.~\ref{fig:vis-swin-resmlp}, \ref{fig:vis-convnext-deit}, and \ref{fig:vis-mixer-swin}. 
In Fig.~\ref{fig:vis-mixer-swin}, the sequence features of the Mixer-B/16 teacher are converted into feature maps for visualization, 
and this conversion results in intermediate layers where the energy concentration is not located at the center. 
This phenomenon reflects the structural characteristics of the teacher model rather than a limitation of UHKD.

\end{document}